\pdfoutput=1

\documentclass[11pt]{article}

\usepackage{acl}

\usepackage[russian,english]{babel}
\usepackage[T1,T2A]{fontenc}  
\usepackage[utf8]{inputenc}   

\usepackage{times}
\usepackage{latexsym}

\usepackage{algorithm}
\usepackage{algorithmic}
\usepackage{amsmath}
\usepackage{bm}

\usepackage[T1]{fontenc}

\usepackage{multirow}
\usepackage{graphicx}
\usepackage{threeparttable}
\usepackage{float}
\usepackage{booktabs} 

\usepackage[utf8]{inputenc}

\usepackage{microtype}

\usepackage{arydshln}

\usepackage{colortbl}

\usepackage{inconsolata}  
\usepackage{graphicx}

\usepackage{CJKutf8}

\title{How Good are LLMs at Relation Extraction under Low-Resource Scenario? Comprehensive Evaluation}

\author{Dawulie Jinensibieke$^1$\thanks{\ \ Equal Contribution.}\ \ , Mieradilijiang Maimaiti$^{1}$\thanks{\ \ Corresponding authors: Mieradilijiang Maimaiti and Xiaobo Wang.}\ \ , Wentao Xiao$^{2*}$, \\
\textbf{Yuanhang Zheng$^3$, Xiaobo Wang$^{1\dagger}$} \\
$^1$XinJiang Technical Institute of Physics and Chemistry Chinese Academy of Sciences \\
$^2$School of Computer Science and Technology, Xinjiang University \\
$^3$Dept. of Comp. Sci. \& Tech., Institute for AI, Tsinghua University, Beijing, China \\
\texttt{dawuliejinensibieke22@mails.ucas.ac.cn; \{miradel,wangxb\}@ms.xjb.ac.cn} \\
\texttt{xiaowentao8@gmail.com; zheng-yh19@mails.tsinghua.edu.cn}
}

\begin{document}
\begin{CJK}{UTF8}{gbsn}

\maketitle
\begin{abstract}
Relation Extraction (RE) serves as a crucial technology for transforming unstructured text into structured information, especially within the framework of Knowledge Graph development. Its importance is emphasized by its essential role in various downstream tasks. Besides the conventional RE methods which are based on neural networks and pre-trained language models, large language models (LLMs) are also utilized in the research field of RE. However, on low-resource languages (LRLs), both conventional RE methods and LLM-based methods perform poorly on RE due to the data scarcity issues.
To this end, this paper constructs low-resource relation extraction datasets in 10 LRLs in three regions (Central Asia, Southeast Asia and Middle East). The corpora are constructed by translating the original publicly available English RE datasets (NYT10, FewRel and CrossRE) using an effective multilingual machine translation. Then, we use the language perplexity (PPL) to filter out the low-quality data from the translated datasets. Finally, we conduct an empirical study and validate the performance of several open-source LLMs on these generated LRL RE datasets.\footnote{The dataset is available at: \url{https://github.com/victor812-hub/entity_datasets}}
\end{abstract}

\section{Introduction}





Relation extraction (RE) aims to discern the semantic relationship between entities within a given context, which serves as a crucial technology for transforming unstructured text into structured information, especially within the framework of Knowledge Graph development~\citep{DBLP:conf/acl/JiG11}. RE also plays an important role in various NLP applications, including schema induction~\citep{DBLP:conf/emnlp/NimishakaviST16} and question answering~\citep{DBLP:conf/acl/YuYHSXZ17}. In previous studies, neural networks and pre-trained language models are widely utilized in the research field of RE~\citep{Aho:DA0029,Aho:DA0030,Aho:DA0031,Aho:DA0032, Aho:DA0033, Aho:DA0034, Aho:DA0035}.

\begin{figure}[t!]
\centering
\hspace{0cm}
\includegraphics[scale=0.42]{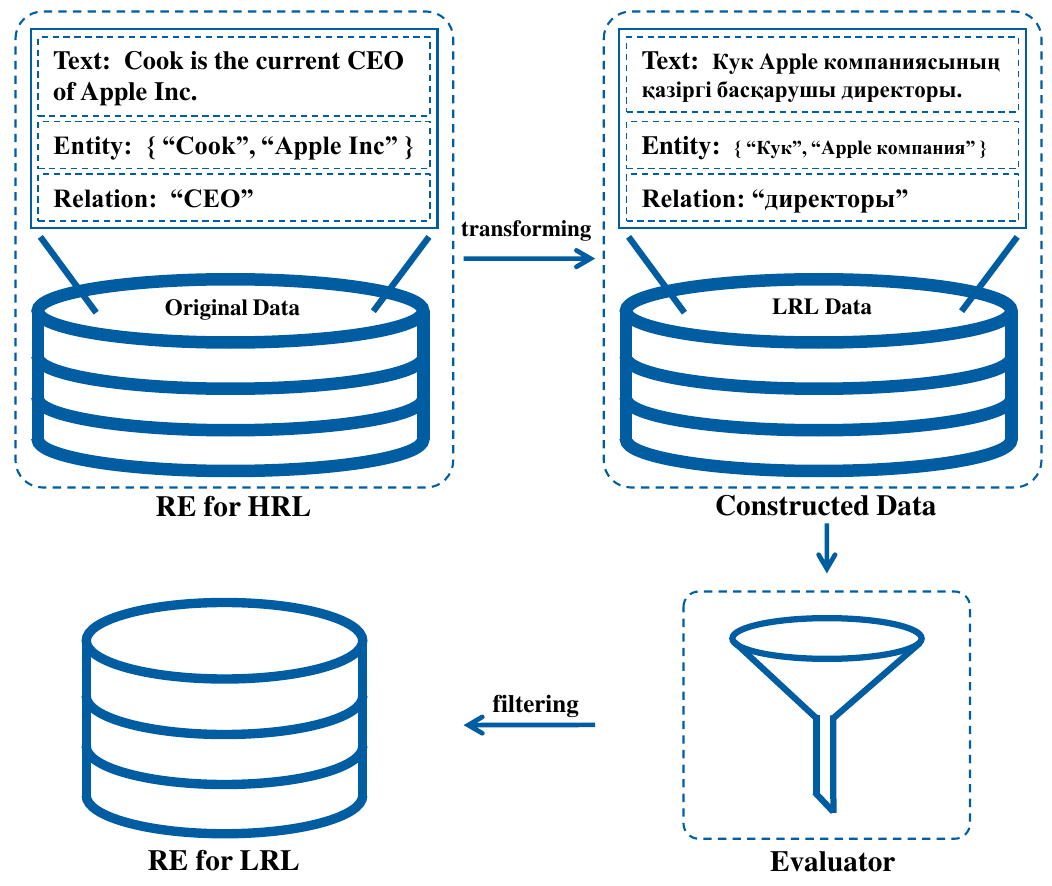}
\caption{The construction process of a low-resource language (LRL) relation extraction (RE) dataset based on a high-resource language (HRL) RE dataset.}
\label{figl}
\end{figure}

\begin{figure*}[t!]
  \centering
  \includegraphics[width=\textwidth]{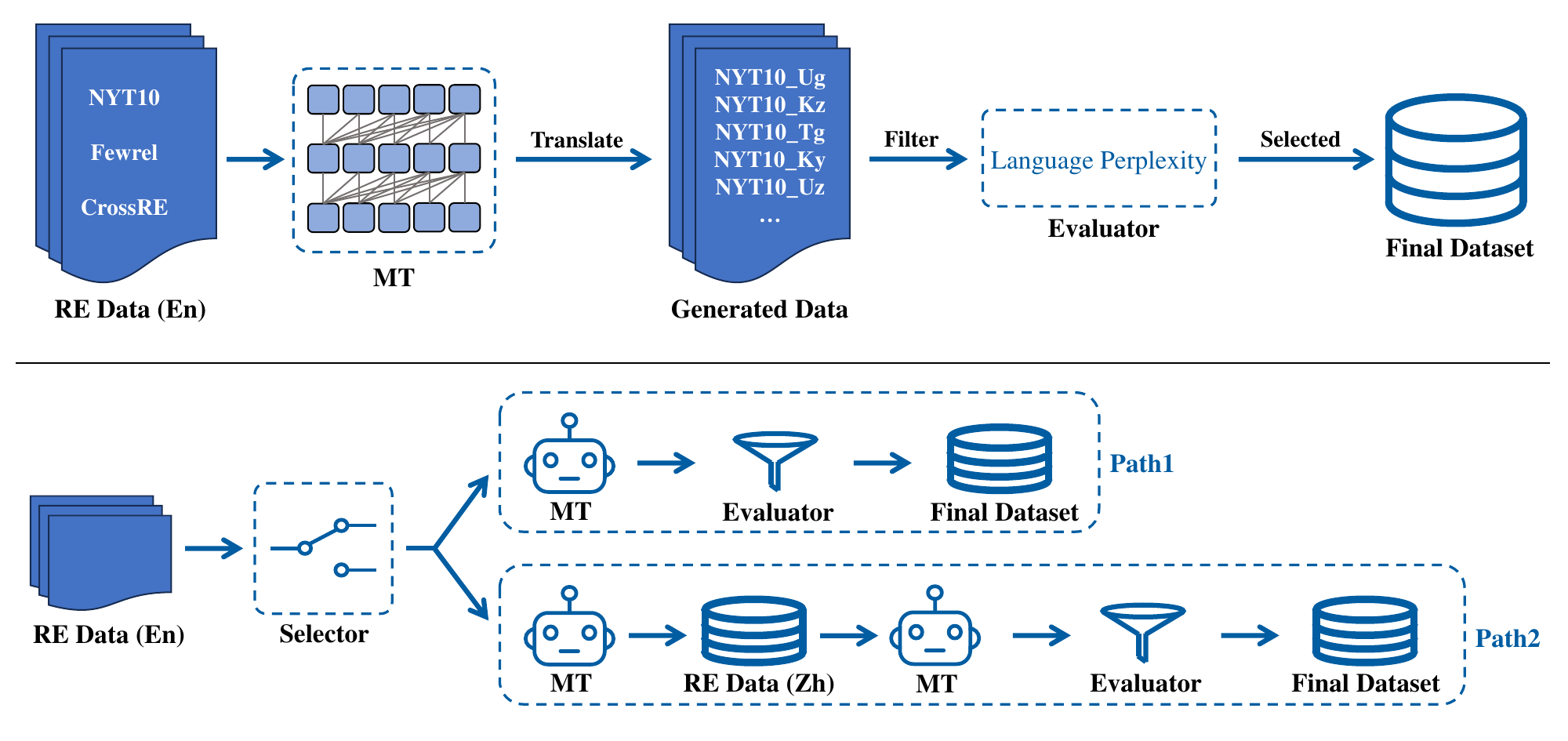} 
  \caption{The detailed process of dataset construction.}
  \label{fig:example}
\end{figure*}

Recently, large language models (LLMs), including GPT-3~\citep{brown2020language}, ChatGPT~\citep{openaichatgptblog} and GPT4~\citep{openaigpt4blog}, have attracted widespread attention within the realm of natural language processing (NLP), since they have exhibited exceptional proficiency in numerous downstream tasks, including machine translation~\citep{moslem-etal-2023-adaptive}, question answering~\citep{kamalloo-etal-2023-evaluating} and information extraction~\citep{kartchner-etal-2023-zero}. In the research field of RE, LLMs have also achieved remarkable performance in various domains~\citep{DBLP:conf/acl/WadhwaAW23,DBLP:conf/eitce/ZhaoCY23,DBLP:conf/hci/AsadaF24}.

Although the previously proposed RE approaches achieve competitive performance on high-resource languages (HRLs), they suffer from extracting the relations between entities under low-resource conditions. On the one hand, the conventional RE methods rely on large-scale labeled data, which are difficult to obtain for low-resource languages (LRLs)~\citep{Aho:DA002}. On the other hand, LLMs also perform poorly on LRLs since there are much fewer pre-training data on LRLs than on HRLs~\citep{DBLP:conf/emnlp/LinMAWCSOGBDPSK22}.


Therefore, in this work, we construct the RE datasets for 10 LRLs from three regions (Central Asia, Southeast Asia and Middle East) by taking advantage of the multilingual machine translation and evaluator model. 
As shown in Figure~\ref{figl}, we first prepare the openly available English RE corpora (NYT10, FewRel, and CrossRE). Then, we exploit a multi-lingual machine translation model NLLB~\citep{Aho:DA0010} to generate the LRL RE data. 
Finally, We further filter out the low-quality data from the translated datasets by evaluating the language perplexity (PPL).
Moreover, we conduct the empirical study and comprehensive evaluations among several open-source LLMs on our constructed corpora.

The main contributions are as follows:
\begin{itemize}
    \item We first constructed the 10 low-resource datasets from three regions: Central Asia, Southeast Asia and Middle East for RE.
    \item We conduct the comprehensive evaluations between LLMs for relation extraction task on our created datasets.
    \item We introduce an efficient evaluation sub-model to increase the quality of generated data effectively.
\end{itemize}

\section{Background}

Relation Extraction (RE) can be conceptualized as a specialized instance of Information Extraction~\citep{Aho:DA0012}. Its primary focus lies in discerning semantic relationships between entities within a given context. The central objective of a typical RE configuration involves extracting meaningful triples from textual data. In this context, a sequence of tokens $T=[t_1, t_2, ..., t_m]$ is considered, a relation set $R=\{r_1, r_2, ..., r_k\}$) and alongside two identified entities represented by spans: $S_A = \{t_i, ..., t_j\}$ and $S_B=\{t_u, ..., t_v\}$. The resulting RE triples conform to the structure $\langle S_A, S_B, r\rangle$, where $r \in R$, and R denotes a predefined set of relation labels. It's noteworthy that due to the inherent directionality of these relations, $\langle S_B, S_A, r\rangle$ denotes a distinct triple capturing a different aspect of the semantic connection.

\begin{table*}
    \centering
    \small
    \begin{threeparttable}
        \begin{tabular}{lccccccccccccc}
            \toprule
            \multirow{2.5}{*}{\textbf{LLMs}} & \multicolumn{8}{c}{\textbf{Central Asia}} & \multicolumn{2}{c}{\textbf{Southeast Asia}} & \multicolumn{2}{c}{\textbf{Middle East}} & \multirow{2.5}{*}{\textbf{Avg.}} \\
            \cmidrule(lr){2-9} \cmidrule(lr){10-11} \cmidrule(lr){12-13}
            & Ug & Kz & Ky & Tg & Tk & Uz & Ug-p2 & Kz-p2 & Fil & Id & He & Fa \\
            \midrule
            LLaMA & 3.0 & \textbf{11.2} & \textbf{13.6} & \textbf{8.4} & \textbf{13.8} & \textbf{13.2} & \textbf{46.2} & \textbf{78.6} & \textbf{13.0} & \textbf{7.9} & \textbf{8.8} & 7.0 & \textbf{18.7} \\
            Falcon & 3.4 & 3.8 & 4.6 & 2.8 & 9.0 & 6.4 & \textbf{46.2} & \textbf{78.6} & 6.2 & 3.4 & 3.4 & 4.2 & 14.3 \\
            BLOOMZ & 0.8 & 4.6 & 3.6 & 5.8 & 2.4 & 3.8 & 5.6 & 5.0 & 5.8 & 1.0 & 1.0 & 4.2 & 3.6 \\
            Qwen & \textbf{3.8} & 7.6 & 12.4 & 7.0 & 9.8 & 7.8 & 4.2 & 6.2 & 7.4 & 7.8 & 7.8 & \textbf{7.4} & 7.4 \\
            OLMo & \textbf{3.8} & 9.4 & 11.4 & 6.6 & 11.2 & 8.6 & 5.0 & 5.0 & 8.8 & 7.2 & 7.2 & 7.0 & 7.6 \\
            \bottomrule
        \end{tabular}
    \end{threeparttable}
    \caption{F1 score of different LLMs on our constructed LRL RE datasets.}
    \label{table_small}
\end{table*}

\section{Methodology}



As shown in Figure~\ref{fig:example}, we first employ a machine translation model to translate the original English RE dataset $\mathcal{D}_{\text{en}} = \{T^{(n)}_{\text{en}},\langle S_A,S_B,r\rangle^{(n)}\}_{n=1}^{N}$ into low-resource language (LRLs) to obtain the LRL RE dataset $\mathcal{D}_{\text{LRL}} = \{T^{(n)}_{\text{LRL}},\langle S_A^{\prime},S_B^{\prime},r\rangle^{(n)}\}_{n=1}^{N}$.
Then, we filter the LRL RE dataset using an evaluator to obtain the final dataset $\mathcal{D}_{\text{final\_LRL}}$.

Specifically, for each sample of the original English RE dataset, we first translate the text $T$ and the entities $S_A$ and $S_B$ into the LRLs, and then we identify the positions of the translated entities within the translated text.

The construction of the LRL RE dataset is facilitated through two distinct pathways, Path1 and Path2. Path1 directly translate the original English RE dataset into LRLs using the machine translation model. Path2 incorporated an additional step of translating into an intermediary language (Chinese) before translating into LRLs, aiming to explore the impact of translation chaining on data quality.

Finally, we use an evaluator to filter the translated dataset to obtain the final LRL RE dataset. Specifically, the filtering process can be divided into two steps. The first step is to remove the samples which are too long or contain illegal characters. The second step is to remove the samples whose text has a high perplexity. Formally, for each sample in LRL RE dataset $\mathcal{D}_{\text{LRL}}$, we calculate the perplexity $\mathrm{PPL}(T^{(n)}_{\text{LRL}})$ of the text $T^{(n)}_{\text{LRL}}$. If the perplexity $\mathrm{PPL}(T^{(n)}_{\text{LRL}})$ is higher than a threshold $\tau$, we remove the sample from the dataset. The final dataset $\mathcal{D}_{\text{final\_LRL}}$ consists of all the remaining samples in $\mathcal{D}_{\text{LRL}}$.


Moreover, we conduct a thorough investigation of existing LLMs on our constructed LRL RE dataset, which aims to provide insights into the adaptability and efficacy of such models in handling LRL scenarios.
In this case, we train the models using our constructed (LRL) RE dataset. The loss function during the training process is defined as:
\begin{equation}
    L(\bm{\theta})=-\log P(r|T^{(n)}_{\text{LRL}},S_A^{\prime},S_B^{\prime};\bm{\theta})
\end{equation}
where $\bm{\theta}$ represents the model parameters.

\section{Experiments}
\subsection{Setup}
\paragraph{Dataset}
Our original relation extraction datasets include NYT10\footnote{\url{https://iesl.cs.umass.edu/riedel/ecml/}}~\citep{Aho:DA0060}, FewRel\footnote{\url{https://huggingface.co/datasets/thunlp/few_rel}}~\citep{Aho:DA0061} and CrossRE\footnote{\url{https://huggingface.co/datasets/DFKI-SLT/cross_re}}~\citep{Aho:DA0062}.
The datasets are translated into the following languages for respective regions: Uyghur (Ug), Kazakh (Kz), Kyrgyz (Ky), Tajik (Tg), Turkmen (Tk) and Uzbek (Uz) for the Central Asian region; for Central Asia; Filipino (Fil) and Indonesian (Id) for Southeast Asia; and Farsi (Fa) and Hebrew (He) for the Middle East region. For Uyghur and Kazakh, we use both Path1 and Path2 to construct the datasets. For other languages, we only use Path1 to build the datasets. We use ``Ug-p2'' and ``Kz-p2'' to represent the Uyghur and Kazakh datasets constructed using Path2, and ``Ug'' and ``Kz'' to represent the Uyghur and Kazakh datasets constructed using Path1.

In the dataset construction process, we employ the NLLB multilingual translation model~\citep{Aho:DA0010} to translate both the text and entities of each data instance from the original dataset. Before the filtering process, the dataset contains approximately 705k samples per language. During the filtering process, we use the multilingual BERT~\citep{devlin-bert} to calculate the perplexity and set the threshold $\tau$ to 20. After the filtering process, the dataset is reduced to approximately 317k samples per language, indicating a filtering out of around 55\% of the data due to quality concerns. The detailed statistics of the datasets are shown in Appendix~\ref{app:stat_dataset}.

\paragraph{Baselines}
We employ our constructed LRLs RE dataset to assess the efficacy of prominent open-source LLMs, gauging their performance in tasks associated with LRL RE. The evaluated models encompass LLaMA\footnote{\url{https://huggingface.co/meta-llama}}~\citep{touvron2023llama}, Falcon\footnote{\url{https://huggingface.co/tiiuae}}~\citep{refinedweb}, BLOOMZ\footnote{\url{https://huggingface.co/bigscience}}~\citep{DBLP:conf/acl/MuennighoffWSRB23}, Qwen\footnote{\url{https://huggingface.co/Qwen}}~\citep{DBLP:journals/corr/abs-2309-16609} and OLMo\footnote{\url{https://huggingface.co/allenai}}~\citep{DBLP:journals/corr/abs-2402-00838}, each of which contains approximately 7 billion parameters.

\begin{figure}[!t]
  \centering
  \includegraphics[width=\columnwidth]{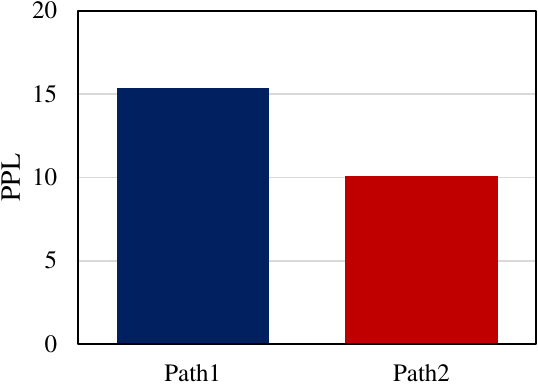} 
  \caption{The perplexity (PPL) value of the Kazakh RE dataset constructed using two distinct pathways.}
  \label{fig:ppl}
\end{figure}







\subsection{Main Results}
As shown in Table~\ref{table_small}, we assess the performance of several large language models (LLMs), including LLaMA, Falcon, BLOOMZ, Qwen and OLMo, on our constructed LRL RE datasets. We evaluate the performance of the models using the F1 score metric. Experimental results show that LLaMA and Falcon showing notably superior performance, which can be attributed to their robust pre-training regimes and their capacity to adapt to the nuances of LRLs.



\subsection{Comparison of Different Pathways}

As shown in Figure~\ref{fig:ppl}, we investigate the effect of different pathways used to construct the datasets. Experimental results show that the dataset constructed using Path2 exhibits a lower PPL value than that of Path1, indicating higher fluency and quality of the dataset constructed using Path2.


\subsection{Effect of Filtering}

As shown in Table~\ref{table_ppl}, we also investigate the effectiveness of the filtering mechanism by calculating the PPL value of the unfiltered and filtered datasets. Experimental results show that the filtering mechanism significantly reduces the PPL, which indicates that the evaluator played a crucial role in refining the datasets obtained from both Path1 and Path2 and there is a need for further refinement to ensure the quality and coherence of the translated texts.



\section{Related Work}

Relation extraction (RE) aims to discern the relationship between two given entities within a contextual milieu. Early RE methods embrace pattern-based techniques~\citep{Aho:DA0027, Aho:DA0028}, alongside CNN/RNN-based approaches~\citep{Aho:DA0029, Aho:DA0030} and graph-based methods~\citep{Aho:DA0031}. With the advent of pre-trained language models (PLMs), a paradigm shift reflects in the widespread adoption of PLMs as the cornerstone for contemporary RE systems~\citep{Aho:DA0032, Aho:DA0033, Aho:DA0034, Aho:DA0035}.


\begin{table}[t!]
    \small
    \centering
    \begin{threeparttable}
        \begin{tabular}{lccc}
            \toprule
            \textbf{Language} & \textbf{Unfiltered} & \textbf{Filtered} & $\bm{\Delta}_{\textbf{PPL}}$ \\
            \midrule
            \textbf{Kz} & 15.37 & 7.27 & 8.10 \\
            \textbf{Ky} & 7.29 & 7.10 & 0.19 \\
            \textbf{Tg} & 9.09 & 8.02 & 1.08 \\
            \textbf{Uz} & 23.99 & 19.50 & 4.49 \\
            \textbf{Id} & 30.96 & 15.41 & 15.55 \\
            \textbf{He} & 11.42 & 9.92 & 1.51 \\
            \textbf{Fa} & 37.05 & 17.41 & 19.64 \\
            \cdashline{1-4}
            \textbf{Kz-p2} & 10.12 & 7.11 & 3.02 \\
            \bottomrule
        \end{tabular}
    \end{threeparttable}
    \caption{Perplexity (PPL) values on unfiltered and filtered datasets for various languages.}
    \label{table_ppl}
\end{table}

Although the aforementioned methods have achieved competitive performance under high-resource conditions, they show limited performance under low-resource conditions. Thus, previous studies have proposed various approaches to improve the models' performance on RE under low-resource conditions. For example, some studies propose harnessing higher-resource data through methods such as Weakly Supervised Augmentation~\citep{najafi-fyshe-2023-weakly}, Multi-lingual Augmentation~\citep{Aho:DA002}, and Auxiliary Knowledge Enhancement~\citep{Aho:DA003}.
Other studies exploit more robust models, employing techniques such as Meta Learning~\citep{obamuyide-johnston-2022-meta}, Transfer Learning~\citep{Aho:DA004}, and Prompt Learning~\citep{Aho:DA005}.
Besides, \citet{Aho:DA0036} propose a simple pipelined approach for entity and relation extraction on HRLs. \citet{Aho:DA0037} present an edge-oriented graph neural model for document-level relation extraction. \citet{Aho:DA0059} propose an optimized method for relation extraction using distant supervision learning.

\section{Conclusion}

In this work, we propose a novel method to generate LRL RE datasets, which translates the original English RE datasets into LRLs and filter out the low-quality samples from the translated datasets. We also evaluate the performance of several open-source LLMs on our constructed LRL RE datasets.

\section*{Limitations}


In this work, we propose a novel method to generate LRL RE datasets. However, this method heavily relies on the machine translation model. If the machine translation model performs poorly when translating the English sentences into the LRLs, the entities in the text may not be correctly translated and thus the LRL RE dataset may not be correctly constructed.

\bibliography{acl_latex}

\clearpage

\appendix

\section{Appendix}

\subsection{Statistics of the Datasets}
\label{app:stat_dataset}

The statistics of the original English RE datasets are shown in Table~\ref{tab:overview-dataset}. The statistics of our constructed LRL RE datasets are shown in Tables~\ref{tab:low-resource-dataset (NYT10)},~\ref{tab:low-resource-dataset (FewRel)} and~\ref{tab:low-resource-dataset (CrossRE)}.

\subsection{Computational Cost for Constructing LRL RE Datasets}


The translation process is conducted on a single NVIDIA GeForce RTX 3090 GPU. For NYT10~\citep{Aho:DA0060}, constructing a RE dataset for a single LRL requires about 5-6 hours. For FewRel~\citep{Aho:DA0061} and CrossRE~\citep{Aho:DA0062}, constructing a RE dataset for a single LRL requires about 1-2 hours.


\subsection{Licenses of Models and Datasets}



The license of the NLLB~\citep{Aho:DA0010} model is CC-BY-NC-4.0. The license of the LLaMA~\citep{touvron2023llama} model is available at \url{https://huggingface.co/huggyllama/llama-7b/blob/main/LICENSE}. The license of the Falcon~\citep{refinedweb} and OLMo~\citep{DBLP:journals/corr/abs-2402-00838} models is Apache-2.0. The license of the BLOOMZ~\citep{DBLP:conf/acl/MuennighoffWSRB23} model is available at \url{https://huggingface.co/spaces/bigscience/license}. The license of the Qwen~\citep{DBLP:journals/corr/abs-2309-16609} model is available at \url{https://github.com/QwenLM/Qwen/blob/main/Tongyi%20Qianwen%20LICENSE%20AGREEMENT}. The license of the FewRel~\citep{Aho:DA0061} dataset is MIT. The license of the NYT10~\citep{Aho:DA0060} and CrossRE~\citep{Aho:DA0062} datasets is not available.

\subsection{Case Study}
To illustrate the practical implications of our dataset construction methodologies, we present a case study involving sample outputs from both Path1 and Path2 in Table~\ref{tab:case_study}. The following table showcases samples of the datasets along with the extracted entities and relations. These examples demonstrate the effectiveness of our dataset construction process in maintaining the integrity and context of the relations despite the complexities involved in translating to LRLs.

\begin{table*}
\centering
\begin{tabular}{lrrrr}
\hline
\textbf{Dataset} & \textbf{Train} & \textbf{Dev} & \textbf{Test} & \textbf{Total}\\
\hline
{NYT10} & {522,611} & {N/A} & {172,448} & {695,059} \\
{FewRel} & {3,200} & {1,521} & {N/A} & {4,721} \\
{CrossRE} & {668} & {2,151} & {2,446} & {5,265} \\
\hline
\end{tabular}
\caption{Statistics of the original English RE datasets (NYT10, FewRel and CrossRE).}
\label{tab:overview-dataset}
\end{table*}

\begin{table*}
\centering
\begin{tabular}{lrrr}
\hline
\textbf{Language} & \textbf{Train} & \textbf{Test} & \textbf{Total}\\
\hline
Ug & 109,686 & 38,573 & 148,259 \\
Kz & 248,290 & 73,578 & 321,868 \\
Ky & 241,724 & 74,238 & 315,962 \\
Tg & 241,752 & 74,976 & 316,728 \\
Tk & 137,627 & 43,799 & 181,426 \\
Uz & 192,851 & 64,646 & 257,497 \\
\cdashline{1-4}
Fil & 270,085 & 81,218 & 351,303 \\
Id & 416,770 & 136,163 & 552,933 \\
\cdashline{1-4}
He & 225,249 & 73,964 & 299,213 \\
Fa & 203,060 & 61,230 & 264,290 \\
\hline
Ug-p2 & 34,251 & 14,924 & 49,175 \\
Kz-p2 & 79,201 & 28,813 & 108,014 \\
\hline
\end{tabular}
\caption{Statistics of the LRL RE datasets derived from the NYT10 dataset.}
\label{tab:low-resource-dataset (NYT10)}
\end{table*}

\begin{table*}
\centering
\begin{tabular}{lrrrr}
\hline
\textbf{Language} & \textbf{Unsup} & \textbf{Wiki} & \textbf{NYT} & \textbf{Pubmed}\\
\hline
Ug & 145 & 85 & 8 & 6 \\
Kz & 310 & 130 & 8 & 15 \\
Ky & 349 & 134 & 7 & 12 \\
Tg & 296 & 55 & 14 & 13 \\
Tk & 338 & 60 & 12 & 8 \\
Uz & 355 & 67 & 17 & 12 \\
\cdashline{1-5}
Fil & 611 & 32 & 26 & 7 \\
Id & 876 & 193 & 46 & 24 \\
\cdashline{1-5}
He & 277 & 173 & 15 & 8 \\
Fa & 482 & 177 & 5 & 15 \\
\hline
Ug-p2 & 37 & 11 & 0 & 0 \\
Kz-p2 & 2 & 4 & 0 & 0 \\
\hline
\end{tabular}
\caption{Statistics of the LRL RE datasets derived from the FewRel dataset.}
\label{tab:low-resource-dataset (FewRel)}
\end{table*}

\begin{table*}
\centering
\begin{tabular}{lrrrrrr}
\hline
\textbf{Language} & \textbf{AI} & \textbf{Literature} & \textbf{Music} & \textbf{News} & \textbf{Politics} & \textbf{Science} \\
\hline
Ug & 57 & 17 & 6 & 411 & 10 & 31 \\
Kz & 60 & 41 & 20 & 442 & 26 & 94 \\
Ky & 72 & 37 & 16 & 463 & 25 & 83 \\
Tg & 79 & 38 & 11 & 472 & 25 & 78 \\
Tk & 65 & 22 & 5 & 460 & 11 & 57 \\
Uz & 96 & 42 & 12 & 491 & 33 & 57 \\
\cdashline{1-7}
Fil & 105 & 36 & 41 & 531 & 38 & 98 \\
Id & 210 & 136 & 63 & 608 & 151 & 223 \\
\cdashline{1-7}
He & 70 & 60 & 11 & 494 & 21 & 56 \\
Fa & 141 & 63 & 21 & 469 & 58 & 93 \\
\hline
Ug-p2 & 41 & 14 & 7 & 370 & 9 & 18 \\
Kz-p2 & 72 & 33 & 19 & 423 & 21 & 69 \\
\hline
\end{tabular}
\caption{Statistics of the LRL RE datasets derived from the CrossRE dataset.}
\label{tab:low-resource-dataset (CrossRE)}
\end{table*}

\begin{table*}
\centering
\begin{tabular}{lp{8cm}ll}
\toprule
\textbf{Path} & \textbf{Text Example} & \textbf{Extracted Entities} & \textbf{Relation} \\
\midrule
Path1 & \foreignlanguage{russian}{Ассоциация өкiлдерi Ксавье Ниельдiң 2002 жылы құрған Free компаниясы туралы айтты; бұл \textcolor{red}{Франциядағы} "Orange" дан кейінгi екіншi iрi интернет-ұйым.} & \{"\foreignlanguage{russian}{\textcolor{red}{Франция}}", "\foreignlanguage{russian}{\textcolor{blue}{Париж}}"\} & "location" \\\midrule
Path2 & \foreignlanguage{russian}{Технологиялық жетiстiктер жайында әңгiме болды; мысалы, 2002 жылы \textcolor{blue}{Парижде} Xavier Niel Free компаниясын құрып, \textcolor{red}{Франциядағы} екiншi iрi интернет-қызметiн бастады.} & \{"\foreignlanguage{russian}{\textcolor{red}{Франция}}", "\foreignlanguage{russian}{\textcolor{blue}{Париж}}"\} & "location" \\
\bottomrule
\end{tabular}
\caption{Sample Outputs from Path1 and Path2.}
\label{tab:case_study}
\end{table*}

\end{CJK}
\end{document}